\documentclass{article}
\usepackage{spconf,amsmath,graphicx}
\usepackage{cite}
\usepackage{amsmath,amssymb,amsfonts}
\usepackage{algorithmic}
\usepackage{graphicx}
\usepackage{textcomp}
\usepackage{subcaption}
\usepackage{tcolorbox}
\usepackage{xcolor}
\usepackage{soul} 
\usepackage{etoolbox}
\usepackage{array, makecell} 
\apptocmd{\thebibliography}{\small}{}{}
\usepackage{makecell} 
\usepackage{subcaption}
\usepackage{afterpage}
\usepackage{multirow}
\usepackage{overpic}
\usepackage{wrapfig}
\usepackage{multirow}
\usepackage{graphicx}
\usepackage{soul}
\usepackage{booktabs}
\usepackage{svg}
\usepackage{tabularx}
\newcommand{\sdrlap}[1]{%
  \rlap{%
    \kern0.15em%
    \textcolor{blue}{%
      \fontsize{2pt}{2pt}\selectfont(#1)%
    }%
  }%
}
\usepackage{booktabs}
\usepackage{colortbl}
\usepackage{multirow}
\usepackage{array}
\usepackage{makecell}   
\usepackage{caption}
\usepackage{tabularx}

\newcommand{\pct}[1]{%
  \rlap{%
    \kern0.12em
    \textcolor{blue}{\fontsize{4pt}{4pt}\selectfont(#1)}%
  }%
}

\usepackage{subcaption}
\usepackage{afterpage}
\usepackage{multirow}
\usepackage{overpic}
\usepackage{wrapfig}
\usepackage{multirow}
\usepackage{graphicx}
\usepackage{soul}
\usepackage{booktabs}
\usepackage{svg}
\usepackage{tabularx}
\usepackage{hyperref}

\usepackage{booktabs}
\usepackage{arydshln}
\definecolor{lightgray}{gray}{0.9}

\setul{0.18ex}{0.15ex}

\title{GameScope: A Multi-Attribute, Multi-Codec Benchmark Dataset for Gaming Video Quality Assessment}

\name{Rajesh Sureddi$^{*}$, Shreshth Saini$^{*}$, Avinab Saha$^{*\mathsection}$, Alan C. Bovik$^{\dagger}$\thanks{$^{\mathsection}$Work done at UT Austin; Avinab Saha is now at Google Research.}}
\address{The University of Texas at Austin$^{*}$, University of Colorado Boulder$^{\dagger}$}
\begin{document}
\maketitle
\begin{abstract}
    The development of video game streaming has grown rapidly, with major platforms such as YouTube and Twitch using different codecs. To support quality assessment models that work consistently across any codec, it is necessary to have access to large, diverse subjective gaming quality datasets. Currently, there are only a few available, each having limitations. To address this gap, we present the largest gaming video quality dataset to date, incorporating both user-generated content (UGC) and professional-generated content (PGC) with extensive visual diversity. Our dataset covers the most widely used codecs—H.264, H.265, and AV1—and consists of 4,048 video samples, each annotated by an average of 37 mean opinion score (MOS) ratings. In addition to overall quality scores, we collect coarse-grained quality attributes, enabling a better understanding of perceptual factors. We study the performance of leading video quality assessment methods on this dataset, including a vision language model that outperforms all the benchmarks. To the best of our knowledge, this is the first dataset that comprehensively addresses gaming video quality assessment across multiple codecs and content types with quality attributes. Our dataset is publicly available at \url{https://rajeshsureddi.github.io/GameScope/}.
\footnote{We thank the Texas Advance Computing Center and the National Science Foundation AI Institute for Foundations of Machine Learning (Grant 2019844) for providing compute resources that contributed to our research.}
\end{abstract}
\section{Introduction}
\label{sec:intro}

Video games are among the most popular forms of entertainment worldwide. The global gaming population is projected to see user penetration rise from 33.57\% to 37.39\%, reaching an estimated 3.04 billion users by 2030 \cite{survey}. Since these gamers are the primary source of streaming content, this projection directly reflects the expanding scale of the game streaming ecosystem.
For streaming platforms, delivering high visual quality is essential to meet viewer expectations. The process of quantifying a video's visual quality is known as video quality assessment (VQA). The most reliable method for VQA is to collect ratings from human subjects, but this approach is time-consuming and does not scale to large datasets. While one could attempt to apply existing non-gaming video quality metrics to gaming videos, prior studies \cite{barman2019no,yu2022perceptual} have shown that gaming content exhibits statistical characteristics distinct from standard video. Consequently, gaming-specific quality metrics must capture these unique features. Furthermore, video games span diverse genres—including Action \& Adventure, Simulation, Role-Playing, Platformer \& Puzzle, Sports \& Racing, Fantasy \& Sci-Fi, and Horror \& Mystery, and user-generated content (UGC) streams often include player faces, overlays, and custom edits that complicate quality assessment. These factors highlight the need for large, diverse datasets covering broad ranges of gaming content to enable robust evaluation of video quality.
As summarized in Table \ref{dataset_h}, several datasets on gaming video quality have been developed. GamingVideoSET \cite{barman2018gamingvideoset} presents an in-lab study on six game contents, extended by KUGVD \cite{barman2019no} to include more distortions. However, these datasets are limited in terms of game variety, resolution, and codec support (restricted to H.264). LIVE-YT-Gaming \cite{yu2023subjective} includes a subjective study on UGC gaming videos, on which the authors developed the GAME-VQP \cite{yu2022perceptual} model. However, it contains a relatively small number of content samples. The LIVE-Meta-MCG database \cite{saha2023study} focuses specifically on mobile gaming quality, on which the authors developed the GAMIVAL model \cite{chen2023gamival}. By contrast, our dataset includes both UGC and PGC, combining mobile and Personal Computer (PC) gaming content with greater variation in genres, three codecs (H.264, H.265, and AV1), and both portrait and landscape resolutions.

In summary, to the best of our knowledge, we propose the largest gaming video quality dataset to date, incorporating both UGC and PGC with extensive variations of game content, genres, and formats. We conducted a comprehensive subjective study on Amazon Mechanical Turk (AMT), obtaining an average of 37 ratings per video, and collected coarse-grained quality attributes to enable perceptual analysis. Furthermore, we benchmark recent state-of-the-art algorithms on this dataset including recent vision-language models.

\begin{table*}[]
\centering
\caption{Summary of existing gaming VQA databases and the proposed GameScope database.}
\label{dataset_h}
\resizebox{\textwidth}{!}{%
\begin{tabular}{lcccccccccccc}
\toprule
\textbf{Database} &
  \textbf{\# Videos} &
  \begin{tabular}[c]{@{}c@{}}\textbf{\# Source} \\ \textbf{Sequences}\end{tabular} &
  \begin{tabular}[c]{@{}c@{}}\textbf{Pristine Source} \\ \textbf{Sequences}\end{tabular} &
  \begin{tabular}[c]{@{}c@{}}\textbf{\# Ratings}\\ \textbf{per Video}\end{tabular}&
  \textbf{Attributes} &
  \textbf{Public} &
  \textbf{Resolution} &
  \textbf{Distortion Type} &
  \textbf{Duration} &
  \textbf{Display Device} &
  \textbf{Display Orientation} &
  \textbf{Study Type} \\ \midrule
GamingVideoSET &
  90 &
  6 &
  Yes &
  25 &
  No &
  Yes &
  \begin{tabular}[c]{@{}c@{}}480p, 720p, \\ 1080p\end{tabular} &
  H.264 &
  30 sec &
  24'' Monitor &
  Landscape &
  Laboratory \\ 
KUGVD &
  90 &
  6 &
  Yes &
  17 &
  No &
  Yes &
  \begin{tabular}[c]{@{}c@{}}480p, 720p, \\ 1080p\end{tabular} &
  H.264 &
  30 sec &
  55'' Monitor &
  Landscape &
  Laboratory \\ 
CGVDS &
  \begin{tabular}[c]{@{}c@{}}360 +\\ anchor stimuli\end{tabular} &
  15 &
  Yes &
  Unavailable &
  No &
  Yes &
  \begin{tabular}[c]{@{}c@{}}480p, 720p, \\ 1080p\end{tabular} &
  H.264 NVENC &
  30 sec &
  24'' Monitor &
  Landscape &
  Laboratory \\ 
TGV &
  1293 &
  150 &
  No &
  Unavailable &
  No &
  No &
  \begin{tabular}[c]{@{}c@{}}480p, 720p, \\ 1080p\end{tabular} &
  \begin{tabular}[c]{@{}c@{}}H.264, H.265,\\  Tencent codec\end{tabular} &
  5 sec &
  \begin{tabular}[c]{@{}c@{}}Unknown\\ Mobile Device\end{tabular} &
  Landscape &
  Laboratory \\ 
\begin{tabular}[c]{@{}l@{}}LIVE-YT-Gaming\end{tabular} &
  600 &
  600 &
  No &
  30 &
  No &
  Yes &
  \begin{tabular}[c]{@{}c@{}}360p, 480p, \\ 720p, 1080p\end{tabular} &
  UGC distortions &
  8-9 sec &
  Multiple Devices &
  Landscape &
  Online \\ 
\begin{tabular}[c]{@{}l@{}}LIVE-Meta \\ Mobile Cloud Gaming\end{tabular} &
  600 &
  30 &
  Yes &
  24 &
  No &
  Yes &
  \begin{tabular}[c]{@{}c@{}}360p, 480p, \\ 540p, 720p\end{tabular} &
  H.264 NVENC &
  20 sec &
  Google Pixel 5 &
  \begin{tabular}[c]{@{}c@{}}Landscape, \\ Portrait\end{tabular} &
  Laboratory \\ 
\begin{tabular}[c]{@{}l@{}}\textbf{GameScope}\end{tabular} &
\textbf{4048} &
\textbf{424} &
\textbf{Yes} &
\textbf{37} &
\textbf{Yes} &
\textbf{Yes} &
\begin{tabular}[c]{@{}c@{}}\textbf{360p, 480p,} \\ \textbf{720p, 1080p,} \\ \textbf{2160p}\end{tabular} & 
\begin{tabular}[c]{@{}c@{}}\textbf{H.264 NVENC,} \\ \textbf{H.265 NVENC, AV1} \end{tabular} &
\textbf{9-10 sec} &
\textbf{Multiple Devices} &
\begin{tabular}[c]{@{}c@{}}\textbf{Landscape,} \\ \textbf{Portrait}\end{tabular} &
\textbf{Online} \\ \bottomrule
\end{tabular}}
\label{tab:dataset-table}
\end{table*}

\section{Related work}

Numerous general-purpose video quality metrics have been proposed in recent years. TLVQM \cite{korhonen2019two} represents one of the earliest models designed for assessing consumer video quality in no-reference (NR) settings. Subsequent approaches, such as RAPIQUE \cite{tu2021efficient}, combine Natural Scene Statistics (NSS) and Convolutional Neural Network (CNN) features, pooling them to train a Support Vector Regressor (SVR) for quality scoring. Similarly, VIDEVAL \cite{tu2021ugc} leverages leading NR-VQA models by extracting and selecting an optimal subset of features to train an evaluative SVR. Moving toward deep learning-based methods, VSFA \cite{li2019quality} incorporates content-aware feature extraction and models temporal memory effects. FAST-VQA \cite{wu2022fast} learns effective video-quality representations via grid mini-patch sampling, aligning local quality fragments to capture global quality efficiently. Patch-VQ \cite{ying2021patch} employs a deep neural architecture that learns to accurately predict both global video quality and local video patch quality. DOVER \cite{wu2023exploring} decouples video assessment into aesthetic and technical perspectives, employing view-specific branches to extract corresponding features and fusing them for a final score. However, the aforementioned methods are often ill-suited for gaming content. To address this gap, GamingVideoSET \cite{barman2018gamingvideoset} and KUGVD \cite{barman2019no} introduced machine learning-based approaches for predicting gaming video quality in no-reference settings. NDNetGaming \cite{NDNetgaming}
combines VMAF-guided pre-training on large-scale data with subjective fine-tuning on gaming content, utilizing temporal pooling to derive video-level quality scores from frame-level predictions.
GAME-VQP \cite{yu2022perceptual} averages predictions from separate SVRs trained on NSS and deep features, while GAMIVAL utilizes a unified SVR on fused NSS and CNN inputs.


Recent advancements in computer vision have significantly facilitated multimodal understanding, particularly between text and vision. Notably, LLaVA-OneVision-1.5 \cite{an2025llava} and Qwen3-VL \cite{Qwen3-VL} propose large-scale vision-language models supporting text, images, and videos. To evaluate the applicability of such foundation models to VQA, we benchmarked our dataset using LLaVA-OneVision-1.5-4B and Qwen3-VL-4B.
Concurrently, several LMM-based quality assessment methods have emerged. Q-ALIGN \cite{wu2023q} aligns visual inputs in a unified pipeline to predict discrete quality levels, which are subsequently converted into numerical MOS. VisualQuality-R1 utilizes reinforcement learning to reduce uncertainty in No-Reference Image Quality Assessment (NR-IQA); although originally designed for images, we employ it for benchmarking via frame-wise prediction. VQAThinker \cite{cao2025vqathinker} also leverages reinforcement learning to jointly model video quality understanding and scoring.

\section{DATASET}
\label{dataset}

\begin{figure*}[t]
    \centering
    \newcommand{\imgwidth}{0.16\textwidth} 
    
    \begin{subfigure}{\imgwidth}
        \includegraphics[width=\linewidth]{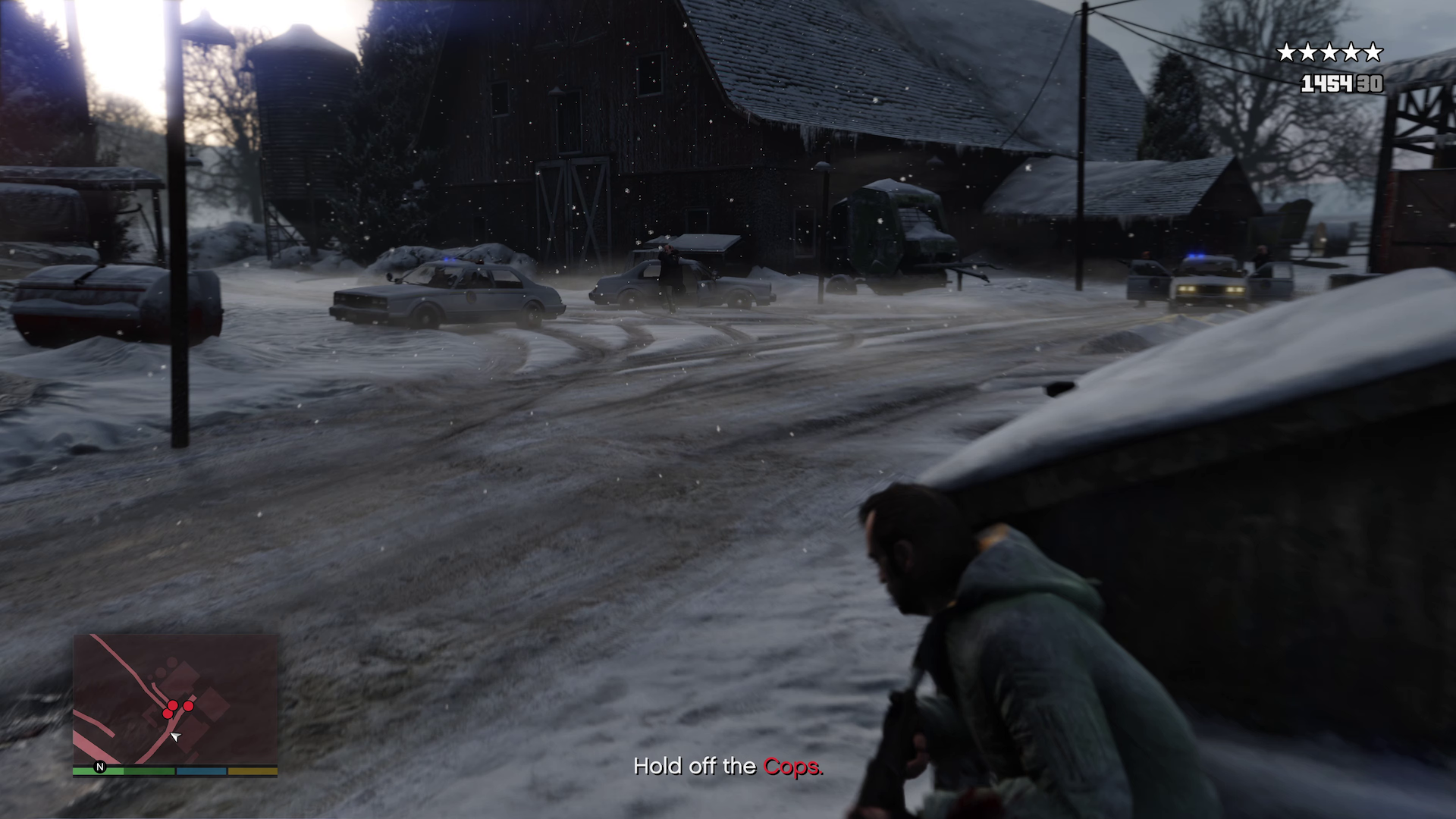}
    \end{subfigure}\hfill
    \begin{subfigure}{\imgwidth}
        \includegraphics[width=\linewidth]{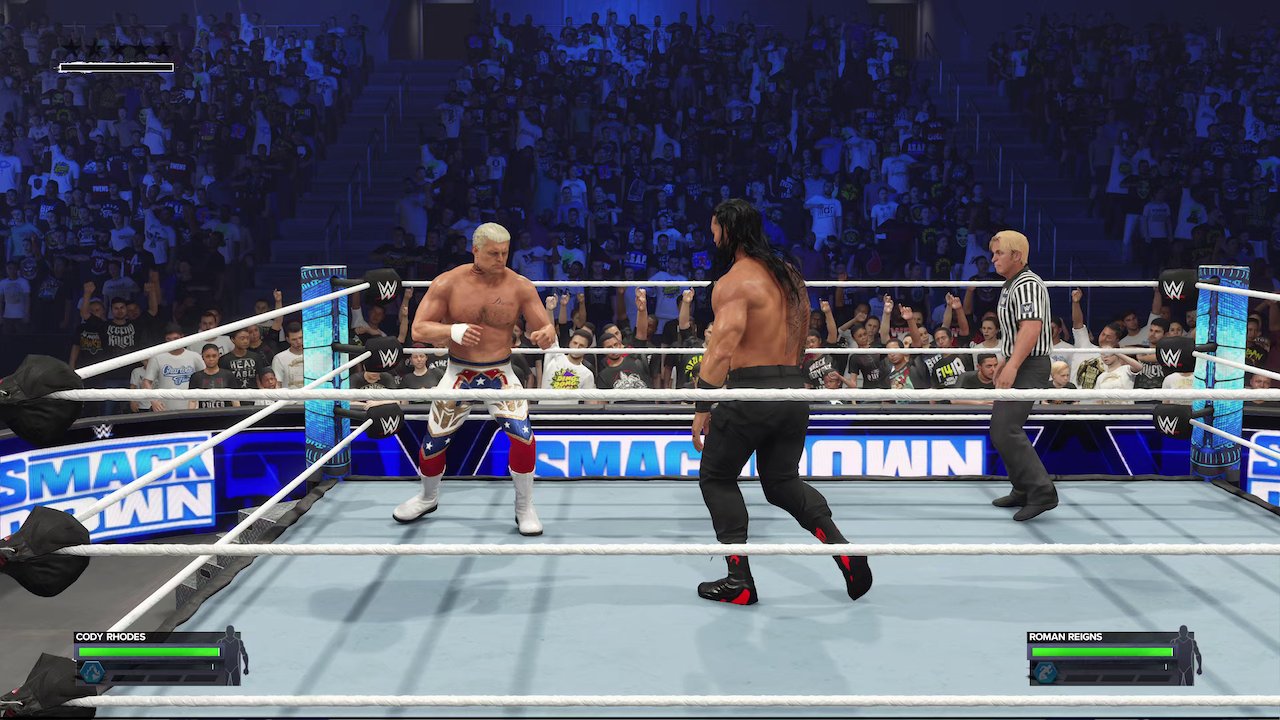}
    \end{subfigure}\hfill
    \begin{subfigure}{\imgwidth}
        \includegraphics[width=\linewidth]{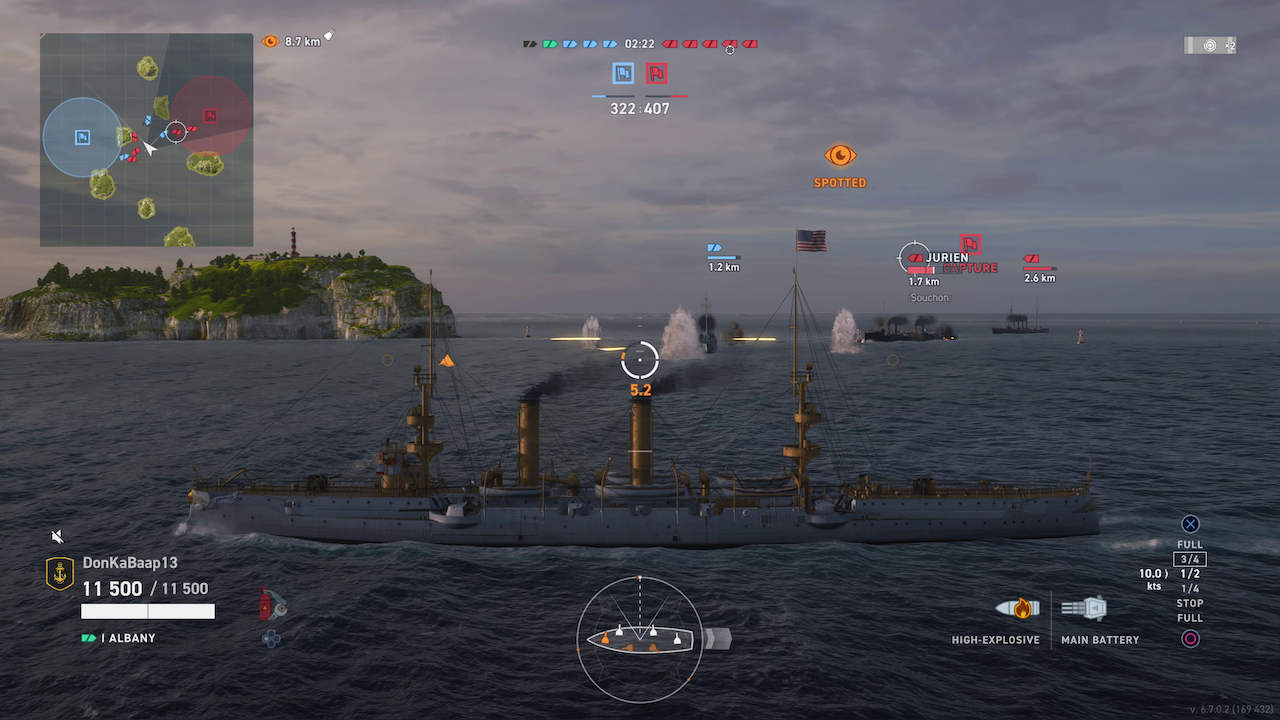}
    \end{subfigure}\hfill
    \begin{subfigure}{\imgwidth}
        \includegraphics[width=\linewidth]{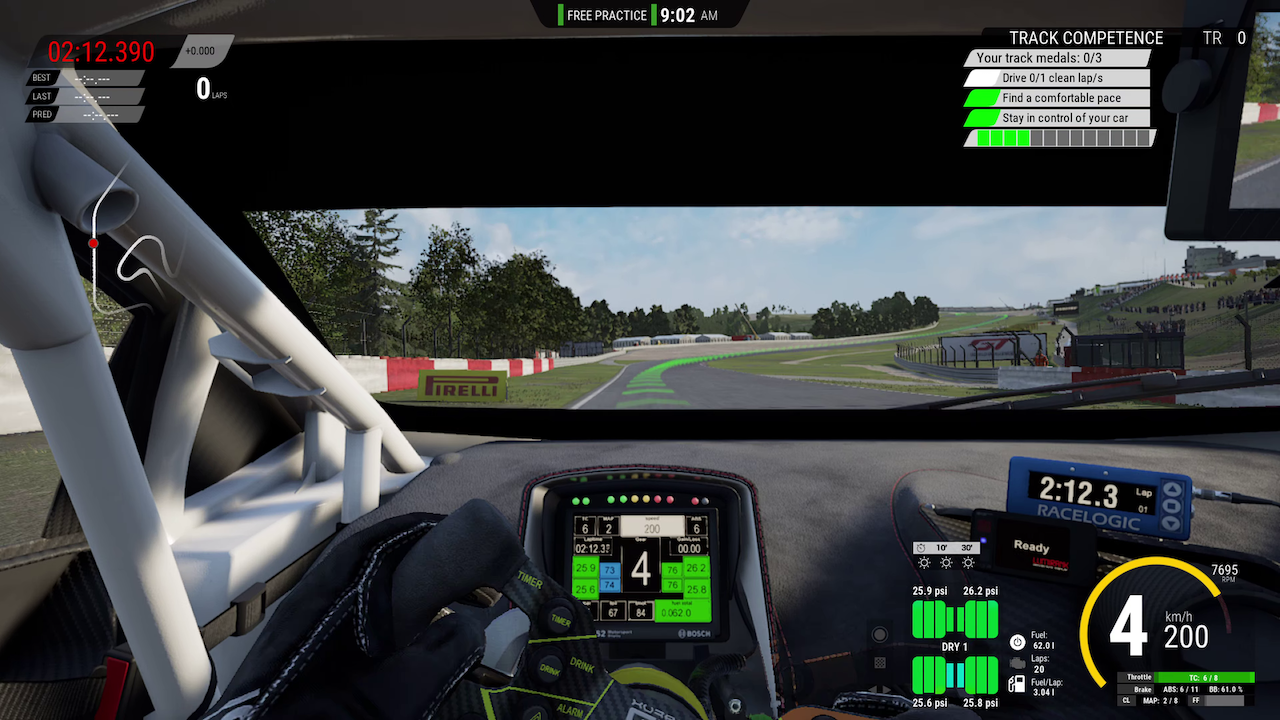}
    \end{subfigure}\hfill
    \begin{subfigure}{\imgwidth}
        \includegraphics[width=\linewidth]{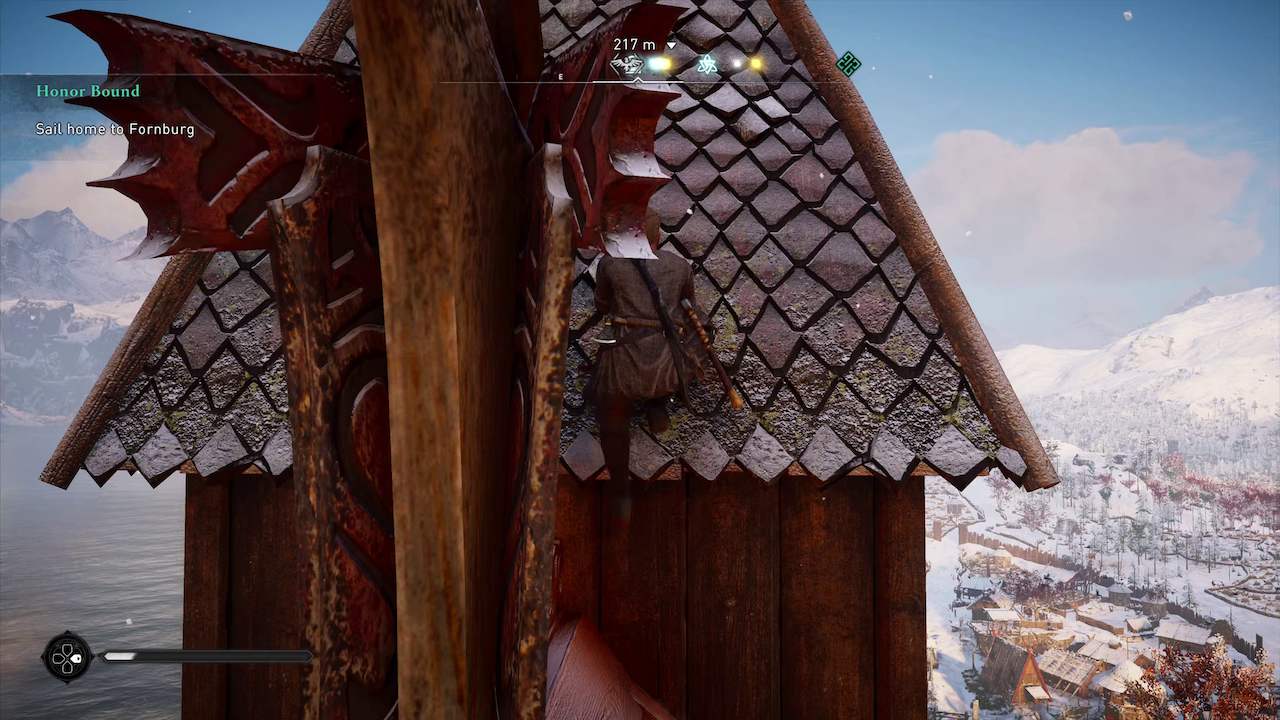}
    \end{subfigure}\hfill
    \begin{subfigure}{\imgwidth}
        \includegraphics[width=\linewidth]{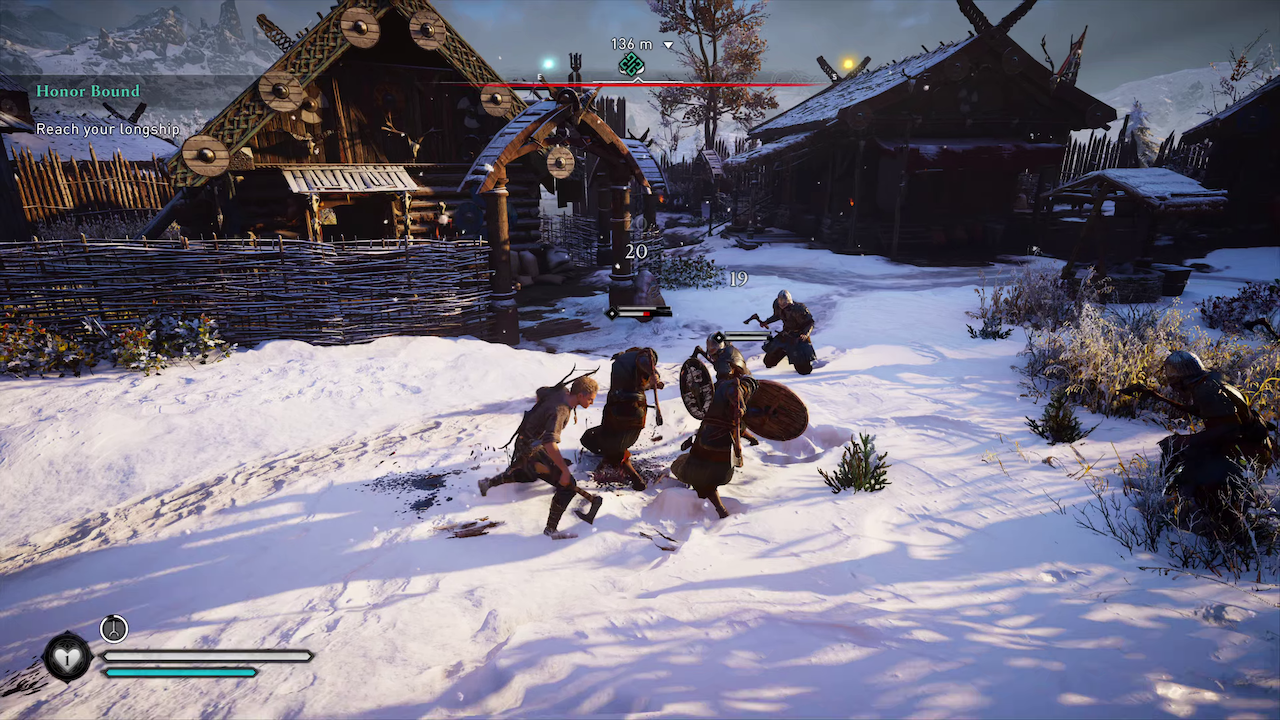}
    \end{subfigure} 

    \vspace{0.5em} 
    
    \begin{subfigure}{\imgwidth}
        \includegraphics[width=\linewidth]{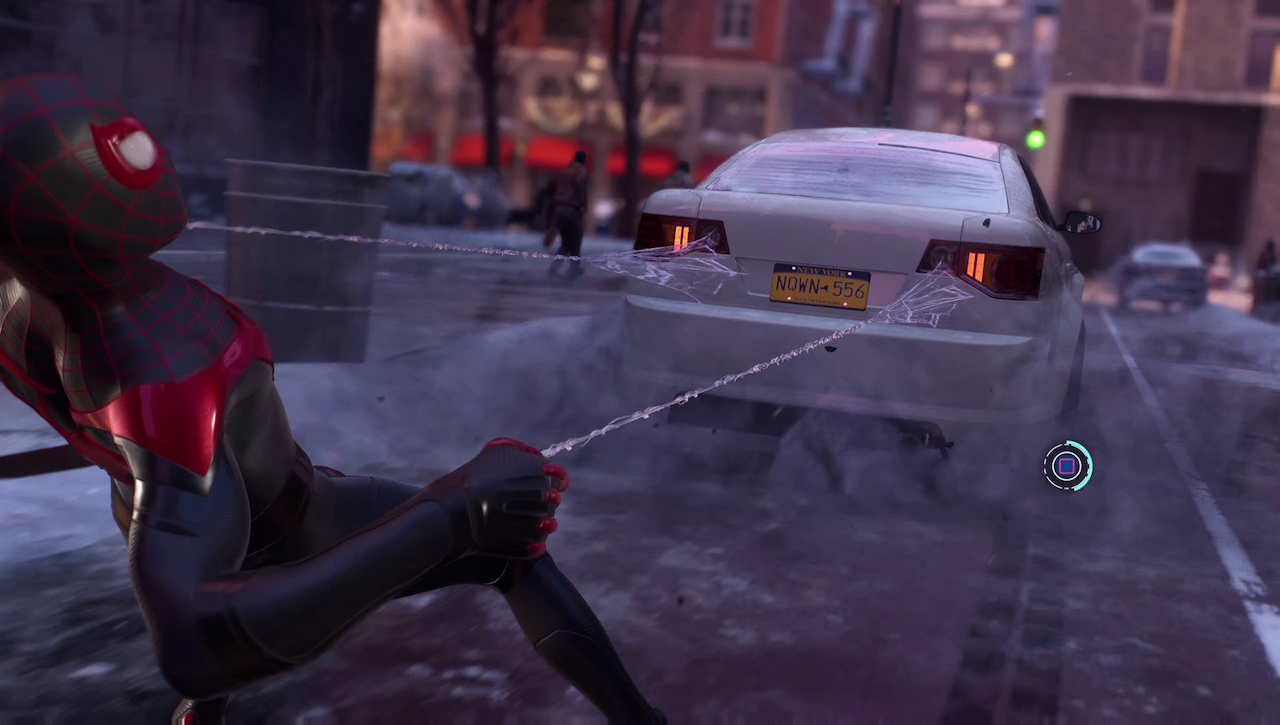}
    \end{subfigure}\hfill
    \begin{subfigure}{\imgwidth}
        \includegraphics[width=\linewidth]{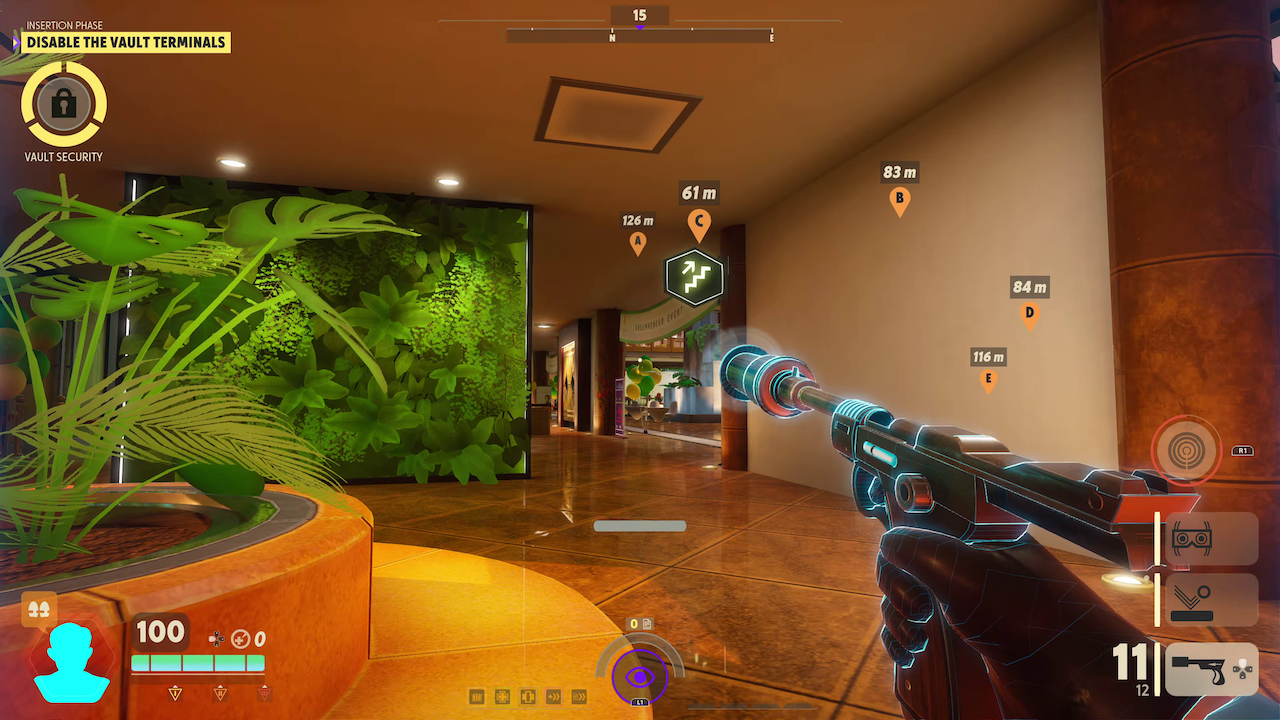}
    \end{subfigure}\hfill
    \begin{subfigure}{\imgwidth}
        \includegraphics[width=\linewidth]{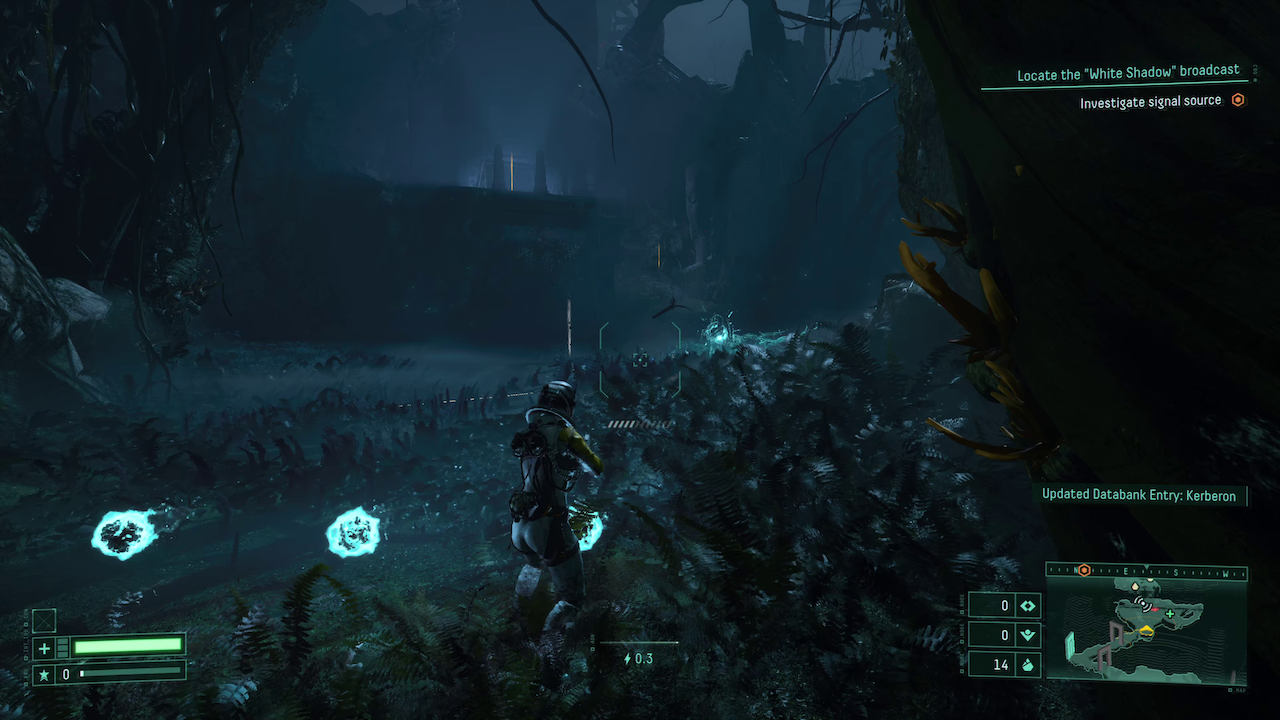}
    \end{subfigure}\hfill
    \begin{subfigure}{\imgwidth}
        \includegraphics[width=\linewidth]{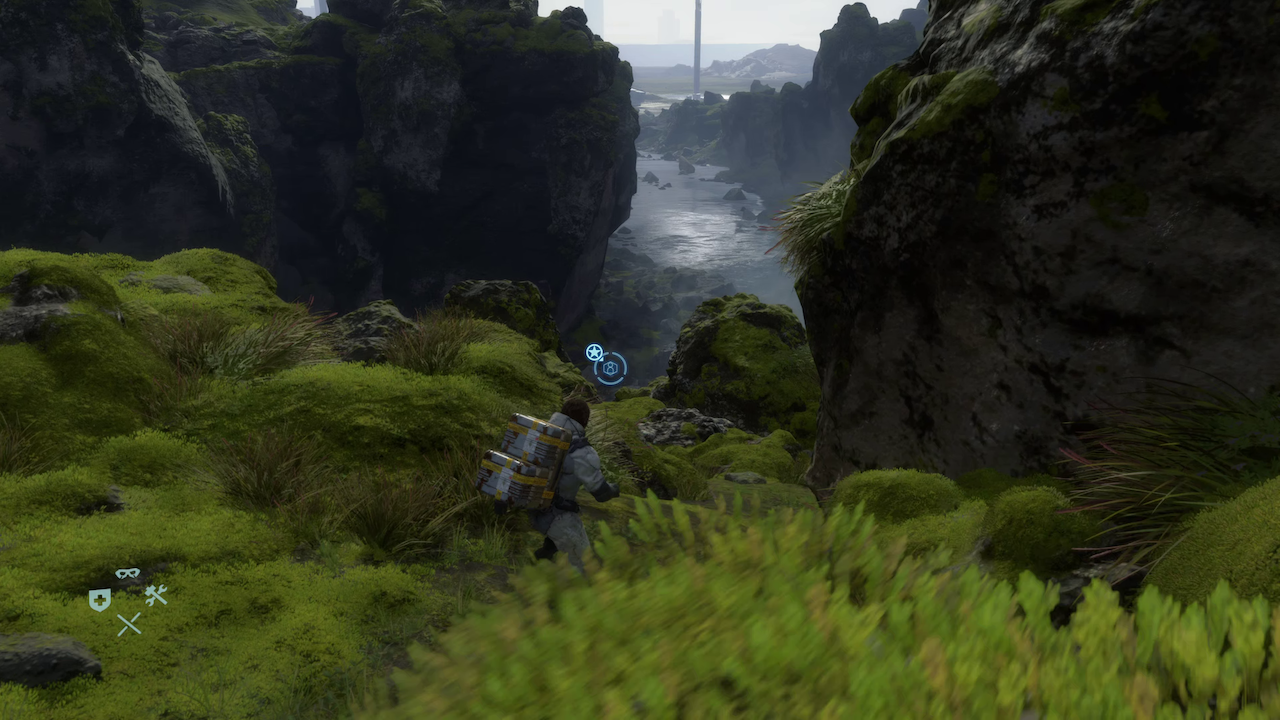}
    \end{subfigure}\hfill
    \begin{subfigure}{\imgwidth}
        \includegraphics[width=\linewidth]{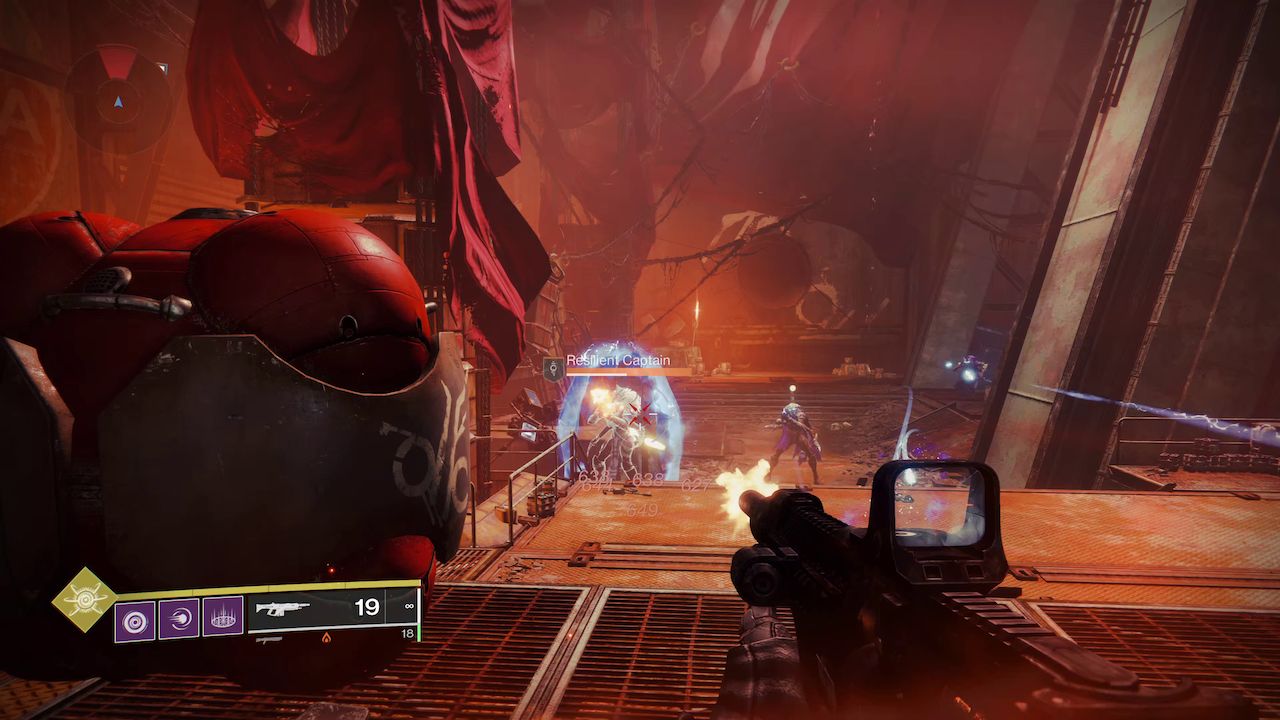}
    \end{subfigure}\hfill
    \begin{subfigure}{\imgwidth}
        \includegraphics[width=\linewidth]{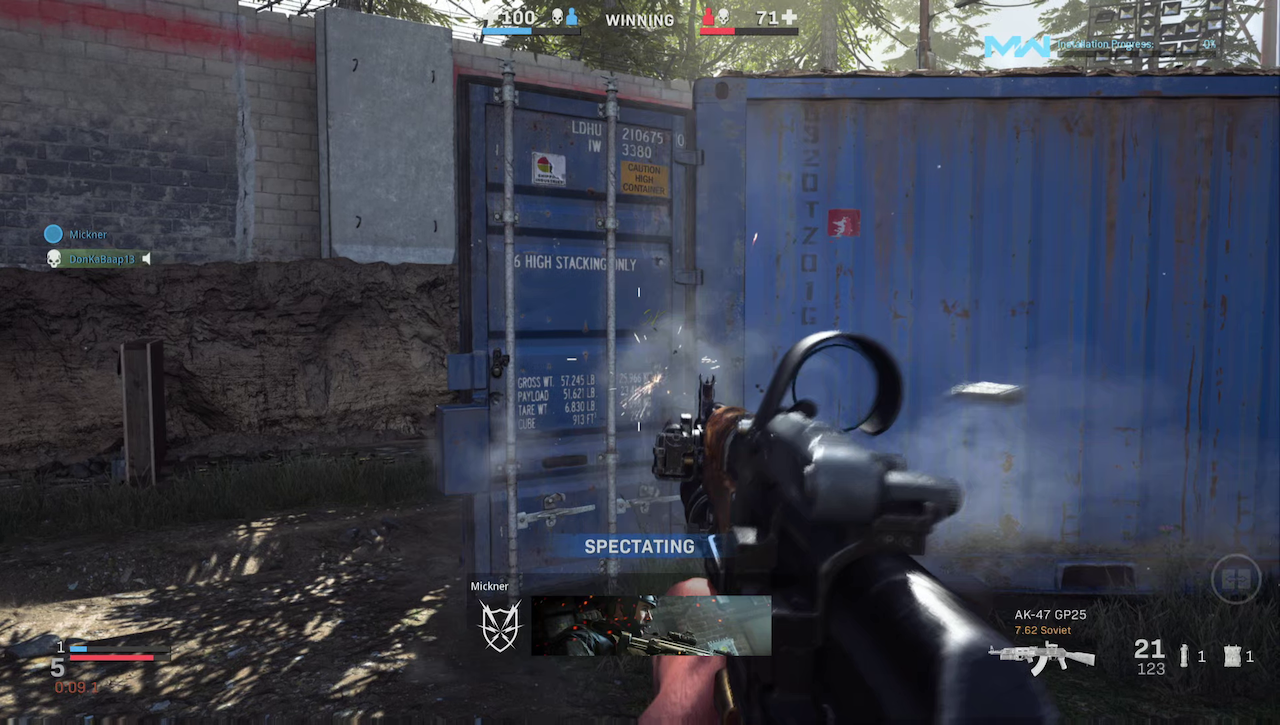}
    \end{subfigure}
    
    \caption{Representative sample frames from the PGC subset of the GameScope dataset, illustrating the diversity of game genres and visual content.}
    \label{fig:12_images}
\end{figure*}
\begin{figure*}[ht]
    \centering
    \begin{minipage}[b]{0.25\textwidth}
        \centering
        \includegraphics[width=\textwidth]{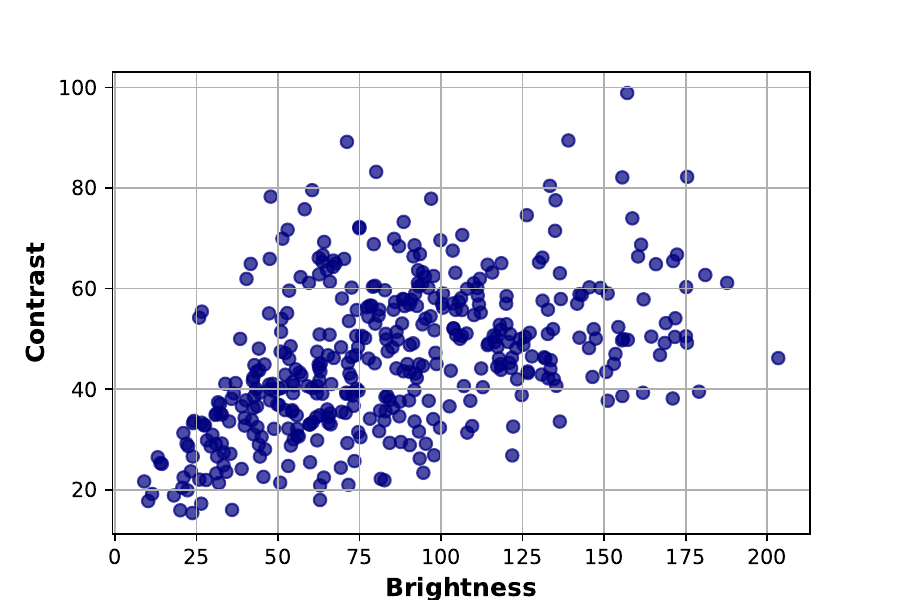}
        \caption*{(a)}
    \end{minipage}
    \hspace{6mm}
    \begin{minipage}[b]{0.25\textwidth}
        \centering
        \includegraphics[width=\textwidth]{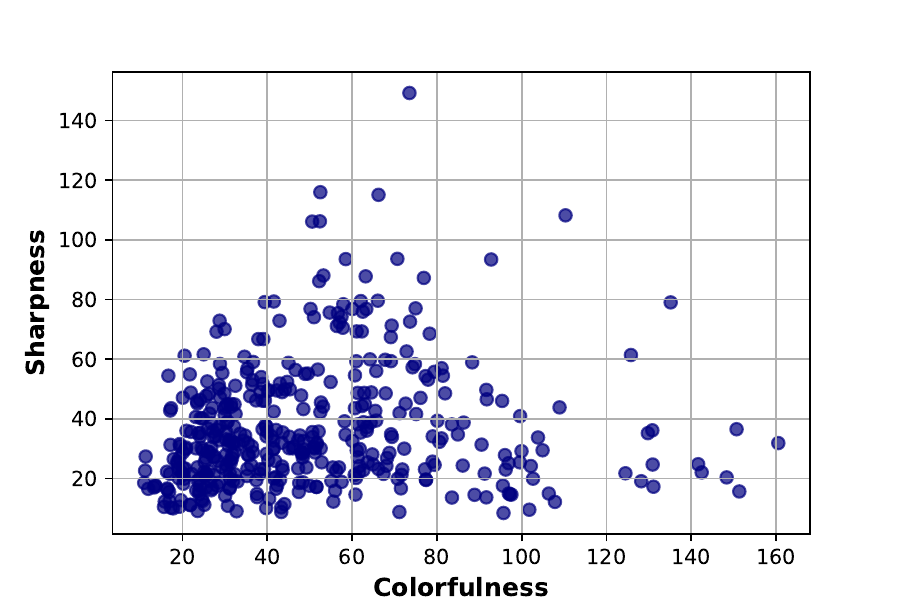}
        \caption*{(b)}
    \end{minipage}
    \hspace{6mm}
    \begin{minipage}[b]{0.25\textwidth}
        \centering
        \includegraphics[width=\textwidth]{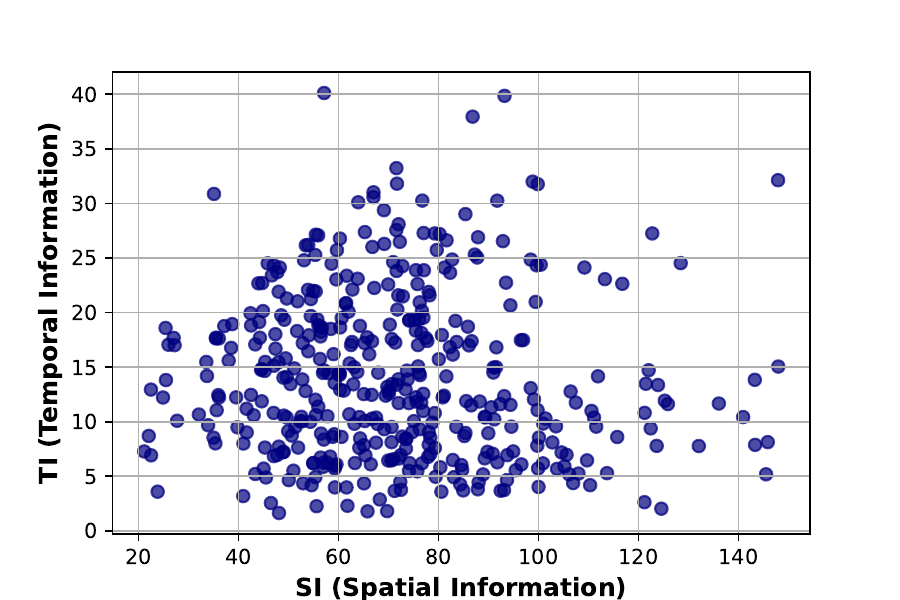}
        \caption*{(c)}
    \end{minipage}
    \caption{The statistics of source content clips (a) Brightness vs Contrast, (b) Colorfullness vs Sharpness and (c) Spatial Information vs Temporal Information.}
    \label{statistics}
\end{figure*}

The compiled dataset comprises 424 source content clips collected from 74 games, each with a duration of ten seconds. The video material was acquired from two distinct sources to represent varying quality tiers. The first subset constitutes User-Generated Content (UGC) covering widely played titles, obtained from YouTube under Creative Commons (CC) licenses. These sequences are characterized by user-applied edits, potentially including overlays such as player faces or logos, and exhibit quality fluctuations resulting from diverse recording software and capture pipelines. The second subset, classified as Professionally-Generated Content (PGC), was captured directly using PlayStation 5 (PS5) internal recording hardware with focusing on recent game releases. Some of the samples are represented in Figure \ref{fig:12_images}. These clips represent direct gameplay rendering untouched by external post-processing or re-encoding stages.
Furthermore, the fundamental characteristics of the source content were analyzed according to ITU-T recommendations \cite{ITU-T_P.910} by measuring brightness, contrast, colorfulness, sharpness, Spatial Information (SI), and Temporal Information (TI). As demonstrated in Figure \ref{statistics}, the quantitative analysis confirms that the collected sequences encompass a broad dynamic range across these spatiotemporal measures, validating the dataset's suitability for this study.

Following data collection, the dataset was prepared by applying a bitrate-resolution ladder to the source sequences. We use three popular encoders H.264, H.265 and AV1 that are widely used in today’s streaming platforms. 
Initially, bitrate recommendations provided by \cite{Twitch_Broadcasting} were evaluated; however, preliminary testing revealed that these values did not yield perceptually distinct quality levels at corresponding resolutions. As we are conducting a controlled study to understand all quality levels, specific bitrate values were manually determined to ensure perceptual separation between levels. The final bitrate ladder utilized is detailed in Table \ref{bitrateladder}. However, as per \cite{Twitch_Broadcasting}, the encoding was performed using Constant Bitrate Mode (CBR) with the ``very fast'' (p1) preset. Additionally, for AV1 encoding specifically, the CPU usage parameter was set to 6. It should be noted that the original source clips were retained alongside the encoded versions to facilitate subsequent subjective assessment. After applying this bit-rate ladder, our final dataset contains 4048 samples that we used in the subjective study.

\begin{table}[h!]
\centering
\small
\caption{Resolution-bitrate (Mbps) ladder employed for each encoder during dataset construction.}
\resizebox{0.4\textwidth}{!}{
\begin{tabular}{lccc}
\toprule
\textbf{Resolution} & \textbf{H.264} & \textbf{H.265} & \textbf{AV1} \\
\midrule
2160p & 5, 8.5, 25 & 1.5, 6, 10 & 1, 5, 10 \\
1080p & 2.3, 5.3, 10 & 0.8, 2, 6 & 0.5, 1.5, 5 \\
720p & 1, 2.5, 5 & 0.6, 1.5, 3.5 & 0.5, 1.3, 3 \\
480p & 1 & 0.8 & 1 \\
\bottomrule
\end{tabular}
}
\label{bitrateladder}
\end{table}

\section{Subjective Study}
To align with the objective of characterizing gaming video quality in streaming scenarios, we conducted an online subjective study via the Amazon Mechanical Turk (AMT) platform. Experimental stimuli were organized into batches using Human Intelligence Tasks (HITs). Each batch was structured to contain a single source sequence alongside its corresponding versions processed by all three evaluated encoders. This grouping strategy was implemented to minimize potential biases arising from content variations, thereby ensuring that subjects focused primarily on assessing quality differences across the encoded conditions. Experimental controls stipulated that a subject could participate in a specific batch only once, though they were permitted to complete multiple distinct batches.

\begin{figure}
    \centering
    \includegraphics[width=0.7\linewidth]{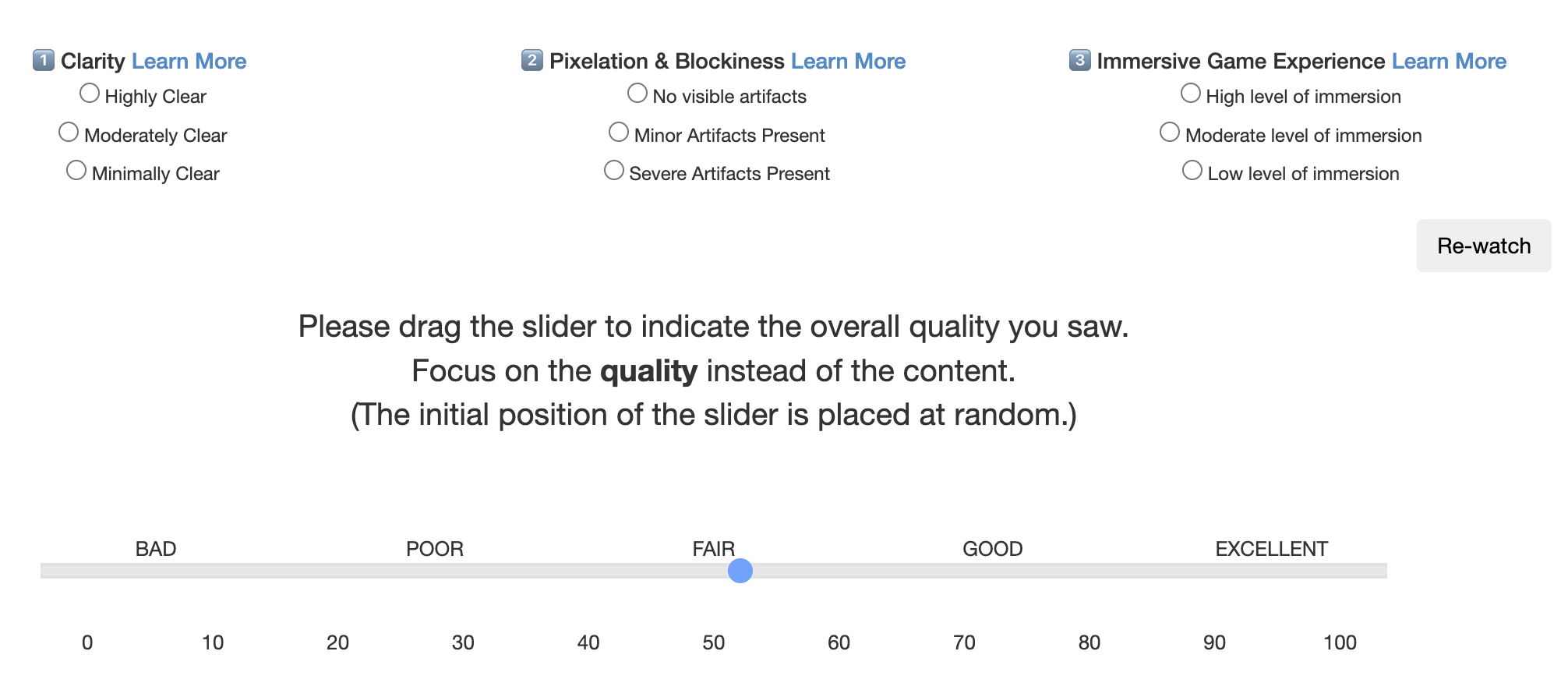}
    \caption{The study template of the subjective quality assessment for collecting quality attributes and overall MOS. Please zoom in for better visibility.}
    \label{rating}
\end{figure}

Data collection proceeded in three phases based on AMT worker qualifications to balance response quality and quantity. In the initial phase, participation was restricted to ``Master" workers possessing a HIT approval rate exceeding 95\% and over 10,000 previous approved HITs; this phase aimed to gather an average of eight responses per sequence. In the second phase, the ``Master" qualification was removed, while maintaining the high approval rate and experience criteria. In the final phase, the qualification threshold was relaxed to a HIT approval rate of greater than 90\%, broadening the participant pool to achieve a minimum target of 30 ratings per stimulus.
The subjective test pipeline was initiated with a comprehensive set of instructions outlining the study's objectives. Participants were presented with exemplar videos representing the full spectrum of standardized quality levels (bad, poor, fair, good, and excellent). To mitigate potential bias, these reference sequences were excluded from the final GameScope dataset. Subjects were explicitly instructed to evaluate technical video quality rather than aesthetic appeal, and the study's ethical guidelines were communicated. Following the instructions, a mandatory quiz verified comprehension; only subjects who successfully completed the quiz were permitted to proceed. The experimental procedure consisted of two distinct phases: training and testing. In the initial training phase, subjects were familiarized with the evaluation interface shown in Figure \ref{rating}. For each stimulus, the video was presented first, followed by the rating tasks. Participants evaluated specific perceptual attributes: Clarity (assessing blur, motion artifacts, and color saturation); Pixelation and Blockiness (capturing encoding-related distortions); and Immersive Game Experience (measuring immersion related specifically to visual quality, independent of game content or genre). To ensure consistent interpretation, examples demonstrating various levels of these attributes were accessible via a ``Learn More" button; forced engagement with these examples was mandatory during training. Finally, subjects provided an overall quality score on a continuous scale from 0 to 100. A ``re-watch" function was available throughout for confirmation. Upon completion of the training phase, the testing phase commenced, during which the experimental ratings were recorded.

\begin{figure}[ht]
    \centering
    \begin{minipage}[b]{0.25\textwidth}
        \centering
        \includegraphics[width=\textwidth]{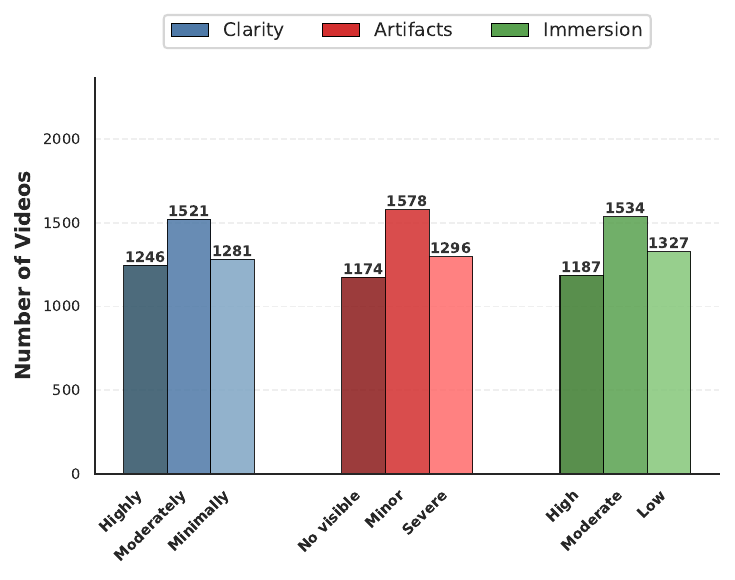}
        \caption{The distribution of quality attributes.}
        \label{fig_att}
    \end{minipage}
\end{figure}
\subsection{Data Screening and Participant Rejection Criteria}
Following data collection, rigorous screening procedures were implemented to ensure data integrity. To guarantee standardized viewing conditions, participation was restricted to traditional computing form factors (PCs or laptops); access via mobile devices or tablets was automatically blocked. Although videos were pre-buffered, data from subjects experiencing more than 40 significant stalls (defined as loading delays $>1$s) were rejected to avoid confounding quality factors. 

Participant reliability was assessed through multiple metrics. Intra-rater consistency was evaluated by presenting six randomly selected videos twice in non-consecutive order; subjects were deemed inconsistent if the difference between repeated ratings exceeded 30 points. Task comprehension was validated using four ``golden videos'' from the LIVE-YT-Gaming dataset \cite{yu2022perceptual} with known MOS. Participants were rejected if ratings deviated from the ground truth by more than 30 points in at least three instances. Furthermore, insincere behaviors (e.g., ``straight-lining'') were identified by analyzing rating distributions; sessions yielding a standard deviation below five were discarded as meaningless. Finally, categorical responses were monitored for repetitive patterns (e.g., consistently selecting the first option). Data displaying such patterns in over 40 instances were excluded due to a lack of genuine engagement.

\begin{table}[ht!]
\definecolor{rowgray}{gray}{0.95}
\definecolor{darkergray}{gray}{0.88}
\setlength{\dashlinegap}{1.5pt}
\centering
\caption{Performance comparison across methods. LLM-based models specifically developed for IQA/VQA are highlighted in gray, and the best results are in bold. }
\resizebox{0.9\linewidth}{!}{
\begin{tabular}{lcccc}
\toprule
\textbf{Method} & \textbf{PLCC} & \textbf{SROCC} & \textbf{KROCC} & \textbf{RMSE} \\
\midrule

\multicolumn{5}{l}{\textbf{SVR Train}} \\
TLVQM          & 0.628  & 0.613  & 0.433  & 0.13  \\
GAME-VQP       & 0.767  & 0.757  & 0.560 & 0.10 \\
RAPIQUE        & 0.727  & 0.694  & 0.511  & 0.11   \\
VIDEVAL        & 0.783  & 0.785  & 0.593  & 0.10    \\
VSFA           & 0.851  & 0.820  & 0.625 & 0.09   \\
GAMIVAL        & 0.856  & 0.851  & 0.659  & 0.08   \\


\multicolumn{5}{l}{\textbf{Zero-shot}} \\
FAST-VQA           & 0.452 & 0.451 & 0.311 & 0.191 \\
DOVER              & 0.520 & 0.524 & 0.368 & 0.173 \\
\rowcolor{rowgray}
VisualQuality-R1   & 0.601 & 0.614 & 0.437 & 0.207 \\ 
\rowcolor{rowgray}
VQA-Thinker-8B   & 0.626 & 0.650 & 0.481 & 0.161 \\

\rowcolor{rowgray}
Q-Align            & 0.711 & 0.715 & 0.533 & 0.152 \\

LLaVA-OneVision-4B & 0.313 &0.275 &0.226&0.189 \\

Qwen3-VL-4B  & \textbf{0.910} & \textbf{0.906} & \textbf{0.784} & \textbf{0.151} \\ 
\bottomrule
\end{tabular}}
\label{tab:results}
\end{table}

\section{Analysis of Subjective Data}

\begin{figure}[ht]
    \centering
    \begin{minipage}[b]{0.3\textwidth}
        \centering
        \includegraphics[width=\textwidth]{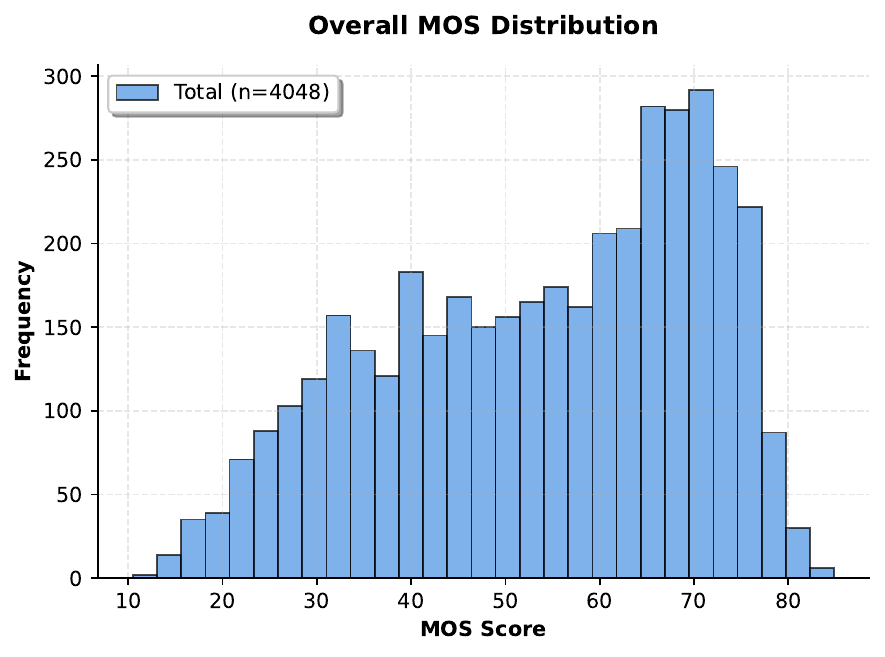}
        \caption{Distribution of collected MOS.}
        \label{mos}
    \end{minipage}
\end{figure}

Using the above subjective rating procedure, we collected 150,874 ratings with an average of 37 per video. To obtain the MOS on each video, a general procedure is to average all raw opinion scores. Specifically, given a video ($v$), and the number N of ratings for that video, with each subject raw score denoted as $r_i$, then
\begin{equation}
MOS{(v)} = \frac{1}{N} \sum_{i=1}^{N} r_i.
\end{equation}
However, the more recent SUREAL \cite{li2020simple} is a more principled and robust model that recovers the ``true" quality score from raw human ratings that can potentially contain noise. They model the observed score ($R_i$) as a combination of true quality ($Q_t$), bias ($b_i$) and subject inconsistency ($\sigma_i$) as follows,

\begin{equation}
R_i = Q_t + b_i + \sigma_i * N(0,1)
\end{equation}
where N(0,1) is a standard normal random variable. they estimate all those parameters from observed raw scores using maximum likelihood. We used the alternating projection method used in \cite{li2020simple}, and computed all the parameters. We visualize the estimated scores in Figure \ref{mos}. One may observe that the MOS distribution is nicely spread over the quality domain, easing the evaluation of existing algorithms and development of new methods. For quality attributes, we chose among the options based on majority vote. As shown in Figure \ref{fig_att}, the sample distortions are spread across all levels, providing good variation of perceptual quality information. To verify rating reliability, we conducted a random split-half analysis using SUREAL over 100 trials. The median PLCC and SROCC between the split groups were 0.92, demonstrating high consistency.

\subsection{Benchmarks}
\begin{figure}[ht]
    \centering
    \begin{minipage}[b]{0.3\textwidth}
        \centering
        \includegraphics[width=\textwidth]{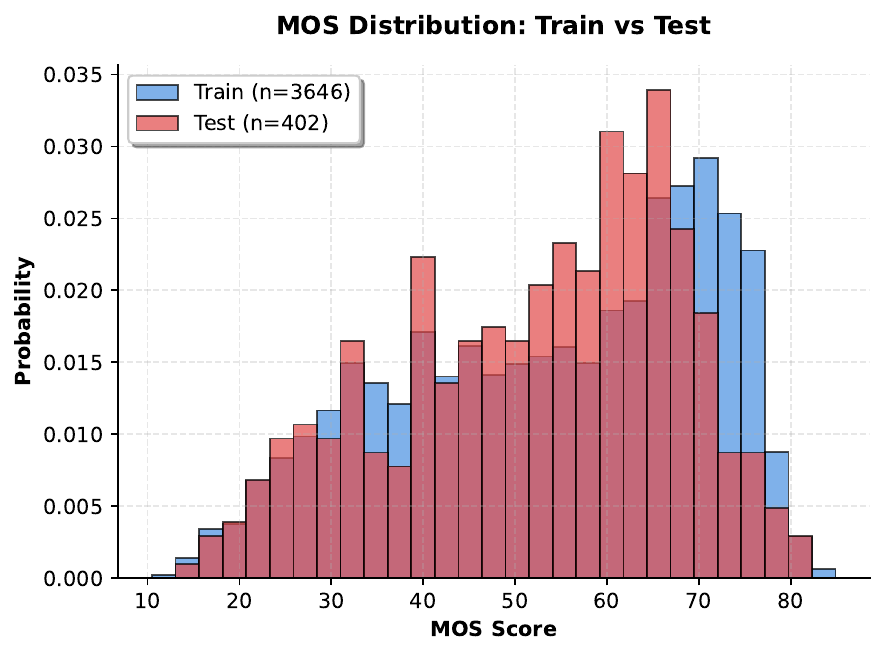}
        \caption{Distribution of the train and test sets against MOS.}
        \label{train_test_mos}
    \end{minipage}
\end{figure}

To ensure consistent benchmarking, we followed the train-test split protocol shown in Figure \ref{train_test_mos}, which strictly separates identical source content to prevent content bias while maintaining proportional distribution of UGC and PGC videos. We recommend this setting to future users of the dataset and will include these standard splits in our public release. During evaluation under this protocol, SVR-based methods were trained exclusively on the training split and evaluated on the test split, whereas zero-shot models were evaluated directly on the test split without any fine-tuning.
As demonstrated in Tables \ref{tab:results} and \ref{tab:text_table}, Qwen3-VL-4B outperformed the existing models in terms of MOS prediction accuracy. It is also competitive for assessing text attributes with LLaVA-OneVision-4B, but both are uniquely capable of predicting both MOS and text attributes in a single inference pass. We provide a naive baseline calculated by always predicting the most frequent class (majority category) observed in the dataset.


\begin{table}[htbp]
    \centering
    \caption{Performance comparison of text quality attribute prediction in terms of accuracy.}
    \label{tab:text_table}
    \begin{tabular}{lccc} 
        \toprule
        \textbf{Method} & \textbf{Clarity} & \textbf{Artifacts} & \textbf{Immersion} \\
        \midrule
        Majority Class & 0.37 & 0.38 & 0.37 \\
        LLaVA-OneVision-4B & \textbf{0.50} & \textbf{0.54} & 0.49 \\
        Qwen3-VL-4B & 0.42 & 0.31 & \textbf{0.98} \\
        \bottomrule
    \end{tabular}
\end{table}

\section{Conclusion}
We introduced the largest gaming video quality dataset to date, comprising both UGC and PGC across three codecs (H.264, H.265, and AV1) with extensive variations in content and resolution. We conducted a large-scale subjective study on AMT to derive reliable Mean Opinion Scores (MOS). Beyond standard scalar ratings, we collected granular quality attributes (clarity, artifacts, and immersion) to facilitate a multidimensional analysis of perceptual quality. Furthermore, we bench-marked leading VQA methods on the new dataset and demonstrated the effectiveness of recent Vision-Language Models at assessing gaming video quality. Our evaluation shows that the Qwen3-VL-4B model delivered superior performance on MOS prediction and was able to compete with LLaVA-OneVision-4B on text attribute assessment. Both are capable of predicting MOS and semantic quality attributes in a single inference pass. Given the diverse distribution of MOS and attributes, we believe this dataset will serve as a foundational resource for evaluating and advancing future gaming VQA algorithms.


\small
\bibliographystyle{IEEEbib}
\bibliography{refs}

\end{document}